%% file: cvpr.tex
\documentclass[final]{cvpr}

\usepackage{times}
\usepackage{epsfig}
\usepackage{graphicx}
\usepackage{amsmath}
\usepackage{amssymb}
\usepackage{booktabs}
\usepackage[linesnumbered,ruled,vlined]{algorithm2e}

\usepackage{comment}
\newcommand{\nop}[1]{}
\usepackage{graphicx}
\usepackage{epstopdf}

\usepackage[pagebackref=true,breaklinks=true,colorlinks,bookmarks=false]{hyperref}

\begin{document}

\title{Learning to Route via Theory-Guided Residual Network}

\author{Chang Liu \\
Shanghai Jiao Tong University \\
Shanghai, China \\
{\tt\small only-changer@sjtu.edu.cn}
\and
Guanjie Zheng\\
The Pennsylvania State University \\
University Park, PA 16802, USA \\
{\tt\small gjz5038@ist.psu.edu}
\and
Zhenhui Li\\
The Pennsylvania State University \\
University Park, PA 16802, USA  \\
{\tt\small jessieli@ist.psu.ed}
}
\newtheorem{problem}{Problem}
\newcommand{\dijtime}{Di-Time\xspace}
\newcommand{\dijdistance}{Di-Dis\xspace}
\newcommand{\astar}{A*\xspace}
\newcommand{\nn}{NN\xspace}
\newcommand{\astarnn}{[A*]NN\xspace}
\newcommand{\lfd}{LfD\xspace}
\newcommand{\query}{Query\xspace}
\newcommand{\ours}{L2R\xspace}

\newcommand{\hangzhou}{Hangzhou\xspace}
\newcommand{\porto}{Porto\xspace}
\newcommand{\jinan}{Jinan\xspace}
\newcommand{\beijing}{Beijing\xspace}

\newcommand{\precision}{Precision\xspace}
\newcommand{\recall}{Recall\xspace}
\newcommand{\fscore}{F1 score\xspace}

\newcommand{\priority}[1]{\textcolor{blue}{[Priority: #1]}}
\newcommand{\prr}{PRR\xspace}
\newcommand{\mscr}{MSCR\xspace}
\newcommand{\rrr}{R3\xspace}
\newcommand{\favour}{FAVOUR\xspace}
\newcommand{\methodFont}{\textit}
\newcommand{\our}{\methodFont{RGM}\xspace}
\maketitle
\input{abstract.tex}
\input{introduction.tex}
\input{related_work.tex}

\input{problem_definition.tex}

\input{method.tex}

\input{experiments.tex}

\input{conclusion.tex}
\input{reference.tex}

\end{document}

%% file: abstract.tex
%
\begin{abstract}

The heavy traffic and related issues have always been concerns for modern cities. With the help of deep learning and reinforcement learning, people have proposed various policies to solve these traffic-related problems, such as smart traffic signal control systems and taxi dispatching systems. People usually validate these policies in a city simulator, since directly applying them in the real city introduces real cost. However, these policies validated in the city simulator may fail in the real city if the simulator is significantly different from the real world. To tackle this problem, we need to build a real-like traffic simulation system. Therefore, in this paper, we propose to learn the human routing model, which is one of the most essential part in the traffic simulator. This problem has two major challenges. First, human routing decisions are determined by multiple factors, besides the common time and distance factor. Second, current historical routes data usually covers just a small portion of vehicles, due to privacy and device availability issues. To address these problems, we propose a theory-guided residual network model, where the theoretical part can emphasize the general principles for human routing decisions (e.g., fastest route), and the residual part can capture drivable condition preferences (e.g., local road or highway). Since the theoretical part is composed of traditional shortest path algorithms that do not need data to train, our residual network can learn human routing models from limited data. We have conducted extensive experiments on multiple real-world datasets to show the superior performance of our model, especially with small data. Besides, we have also illustrated why our model is better at recovering real routes through case studies.

%

\end{abstract}

%% file: introduction.tex
\section{Introduction}
\label{sec:introduction}


The rapid urbanization process has created many massive cities where millions of people live in. Every day, people may travel around the city for work, entertainment, and food. It has stimulated a large volume of traffic traveling around the city and caused many urban problems, e.g., traffic congestion, noise, and air pollution. People have been trying to use machine learning technologies to generate some policies to alleviate these problems. However, they can not directly apply these policies in the real world since they will generate unacceptable real costs (e.g., traffic congestion, traffic accidents). 

To verify these transportation policies feasibly, the solution is to build a city traffic simulator, in which people can easily train and test their policies. This simulator can significantly help people project the traffic under unseen cases (e.g., concerts ), develop effective management policies, validate city construction plans, and even plan a new city. However, in order to make sure these policies can perform well after transferring from simulator to the real world, \textbf{the traffic simulator has to be close enough to the real world}, which means the simulated traffic is similar to the real-world traffic. Generally, the key to building such a real-like traffic simulator is the accurate estimation of the following two inputs: (1) OD (origin-destination): how many people travel from one area to another; (2) Routing: what routes people take to their destinations.

\nop{The recent development of sensors and devices have enabled vast amount of human mobility data to be collected. Building upon these data, many kinds of urban applications, like traffic volume estimation and speed prediction can be further developed. Everyday, people need to travel in the cities to work, eat and entertain. In order to understand how people move in the cities, we need to infer what routes people will choose when they travel to their destinations. This way, the real-time traffic volume at each road segments in cities can be modeled, which forms base of further applications like event control, city function management, and infrastructure planing.}

In this paper, we propose to learn how people choose their routes from mobility data (e.g., driving trajectories). Given the current traffic situation (e.g., road network, traffic volume, and speed) and people's OD, we aim to inference the route they will choose. We aim to learn the real decision-making process when people plan their routes. 

The setting of learning to route problem is similar to imitation learning, especially behavior cloning~\cite{Sammut2010}, which aims to learn and mimic human behavior in the real world. Therefore, our learning to route problem is different from traditional route recommendations and navigation systems.  Generally, these studies~\cite{shimizu1995route, nakajima2012route} try to find routes with specific requirements given the current road network and traffic condition. However, the recommended routes might be very far from the real routes that people have chosen. Statistics from existing datasets~\cite{lian2018one} show that \textbf{only 50\% of the drivers follow the recommended path by the navigation systems}. It means there are other factors affecting people's routing decisions, and that is what we really want to learn in our learning to route problem. 

In fact, the multiple factors affecting people's routing decisions have made it a challenging problem. Although time and distance are the primary considerations when people choose routes, their decisions may also get affected by drivable conditions, like the level of roads, the width of roads, and the number of turns.




In addition, the routing trajectory data in the real world is usually very limited and sparse. Usually, only trajectories of a certain portion of all the vehicles will be collected. For instance, in a public one-year Porto taxi route dataset~\cite{moreira2013predicting}, there are in total more than 100 road segments and 5,000 possible routes in the downtown area of Porto. However, we only have 2576 records covering this region for the whole year. Considering the varying traffic situation, this amount is significantly insufficient to train an accurate pure data-driven model. 

Based on the previous facts, we propose a theory-guided data-driven model to learning human routing decisions from data, in which the theoretical part can address the general principles (e.g., Dijkstra and A*) and the data-driven part can incorporate drivable conditions. Specifically, inspired by the success of the residual learning~\cite{he2016deep} and the boosting idea~\cite{chen2016xgboost} in conventional supervised learning tasks, we propose to use a residual framework to combine the theoretical part and the data-driven part. We have also shown its superior performance over other frameworks, especially when the data amount is insufficient. In summary, our contributions can be summarized as follows:

\begin{itemize}
    \item We are the first to raise the learning to route problem. This problem is different from the traditional route recommendation problem, which aims at recommending the routes with specific standards (e.g., fastest).
    \item To address the learning to route problem, we propose a residual learning model that combines the theoretical model and the deep neural network, which can resolve the sparsity issue in the route data.
    \item We conduct experiments on multiple real-world datasets to verify the effectiveness of the proposed method. We have also demonstrated that the learned routes can capture the multiple aforementioned factors through a case study. 
\end{itemize}


%% file: related_work.tex
\section{Related Work}
\label{sec:related-work}

\subsection{Route Recommendation}

Route recommendation is the most relevant topic to our paper, which aims to recommend routes for a given origin and destination that can save time for travelers or mitigate traffic congestion. Tradition navigation systems and map servers usually first estimate the mean travel time evaluation on each link, then utilize Dijkstra based method\cite{shimizu1995route}, A* based method\cite{bell2009hyperstar}, or dynamic programming~\cite{Weymann1995OptimizationOT} on a graph with edges weighted by distance, time, or fuel consumption. 

A common way to consider all factors is to build a comprehensive graph with multiple factors~\cite{ hamada2014route, Guo2018LearningTR, wen2017efficient, li2015exploring}. People have proposed many approaches to process such a comprehensive graph: \cite{dai2015personalized,su2014crowdplanner} generate candidates routes from historical data and chooses one from candidates, which can choose routes according to the specific requirements of users.
Some researchers~\cite{Ghagage2018ASO} begin to learn users' preferences by the historical data: \cite{campigotto2016personalized} first predicts users’ preferences, then use the learned preferences to recommend routes by linear regression. \cite{dong2014recommend} assigns scores on different roads based on historical data. Some researchers~\cite{hu2012pick,wang2014r3,dong2014recommend} aims at recommending routes to taxis, to minimize congestion while maximizing profits. \cite{Guo2018LearningTR} utilizes transfer learning to alleviate insufficient data problem.

The above papers all focused on finding the routes that fit certain standards, while \cite{holleczek2015traffic} shows that those recommended routes are always different from the ones that people take in the real world. In this paper, we propose to solve a new problem. We aim to model the human routing decisions from multi-source data, and retrieve routes that are close to the real routes planed by humans. 

Indeed, some aforementioned works try to learn users' preferences\cite{dai2015personalized,su2014crowdplanner}, which is similar to our goal. However, most preferences learned by their method are the statistic based on the historical routes, while historical routes are usually hard to acquire. Instead, we also take real-time drivable condition on each road and intersections into consideration, which enable our method to better learn the real decision-making process for routing. We will compare our proposed method with their methods in our experiments. The results show that the users' preferences learned by their methods are limited to address our learning to route problem. 

Besides, some people tend to re-design their routes during the journey if real-time traffic has changed a lot~\cite{Feng2016ASO, Ghagage2018ASO}. This issue exists both in our problem and nowadays navigation systems. The common way is to re-query to the map servers every once in a while. In our problem, we can also solve this issue by re-querying to our model every several time intervals.

\subsection{Utilizing Theoretical Knowledge}

Recently, due to the complex mechanisms behind the real-world problems and the limited data we have, researchers start trying to add theoretical prior knowledge into the data-driven learning process~\cite{karpatne2017theory, wagner2016theory}. They have come up different methodologies in doing so, including theory-guided model design~\cite{twa2005automated}, adding theories as regularization, constraints or probabilistic prior~\cite{schrodt2015bhpmf, karpatne2016global},  using theoretical knowledge to do preprocessing or postprocessing~\cite{fischer2006predicting}, hybrid theory-data models~\cite{sadowski2016synergies, wang2017physics}, and using data to augment theoretical models~\cite{evensen2009data, chapelle2011empirical}. We refer the interested readers to the survey~\cite{karpatne2017theory}. Other than other categories of methods which only apply when theoretical knowledge exist in a certain form, we propose a general hybrid theory-data model. Specifically, we utilize the residual structure~\cite{he2016deep, ajay2019combining} to combine the data-driven part and theoretical part, which allows the data-driven model to benefit from the theoretical knowledge and improve upon it.

%% file: problem_definition.tex
\section{Problem Definition}
\label{sec:problem-definition}

Before introducing the problem, we first summarize the key notations in Table~\ref{tab:notation}.

\begin{table}[!htbp]
\centering
  \caption{Notations}
  \label{tab:notation}
  \begin{tabular}{ll}
    \toprule
    Notation&Meaning\\
    \midrule
    $i$ & Index of route  \\
    $j$ & Index of road  \\
    $G$ & Road network \\
    $M$ & Number of roads \\
	$N$ & Number of intersections \\
	$R_{i}$ & $M \times 1$ binary vector denotes $i$-th route\\ 
	$S_{i}$ & Start point of $i$-th route\\
	$E_{i}$ & End point of $i$-th route\\
	$T_{i}$ & Departure time of $i$-th route\\
	$A_{i}$ & $\left\{S_{i},  T_{i}, E_{i}\right\}$\\
	$V_{j,t}$ & Average speed in $t$-th time interval on $j$-th road\\
	$C_{j}$ & Drivable condition on $j$-th road\\
	 \bottomrule
\end{tabular}
\end{table}

To learn human routing decisions, we need to use the following multi-source data, including road network data, route data, real-time traffic data, and road condition data. \nop{We use bold letter $\mathbf{B}$ to represent a set $B$.}
\begin{itemize}
    \item \textbf{Road network data} defines a graph $G$ with $N$ nodes (intersections) and $M$ edges (roads), which is abstracted from the real-world city.
    \item \textbf{Route data} refers to the historical routes in $G$, defined by $R = \left\{R_{1}, R_{2}, \cdots, R_{K}\right\}$, where $K$ is the number of routes. Each route $R_{i}$ is an $M$-dimension binary vector indicating whether each road is included. Each route also has extra attributes $A_{i}$, including start point $S_{i}$, departure time $T_{i}$, and end point $E_{i}$.
    \item \textbf{Real time traffic data} is defined by $V_{j,t}$, which stands for the average vehicle speed on $j$-th road in $t$-th time interval. 
    \item \textbf{Drivable condition data} is defined by $C_{j}$, which is a feature matrix describing the static road conditions and people's preference on $j$-th road, like the number of drivable lanes, speed limit and road rank.
\end{itemize}

\nop{
\begin{table}[!htbp]
\centering
  \caption{Notations}
  \label{tab:notation}
  \begin{tabular}{ll}
    \toprule
    Notation&Meaning\\
    \midrule
    $i$ & Index of route  \\ \hline
    $j$ & Index of road  \\ \hline
    $G$ & Road network \\ \hline
    $M$ & Number of roads \\ \hline
	$N$ & Number of intersections \\ \hline
	$R_{i}$ & $i$-th route\\ \hline
	$A_{i} = \left\{S_{i},  T_{i}, E_{i}\right\}$ & Attributes of route $i$\\
	& $S_{i}, E_i, T_i$ for start point, end point \\ \hline
	& and departure time correspondingly \\
	$V_{j,t}$ & Average speed in $t$-th time stamp \\
	          & on $j$-th road \\ \hline
	$C_{j}$ & Driving condition on $j$-th road\\ 
	 \bottomrule
\end{tabular}
\end{table}
}

Our problem is formulated as follows.

\nop{Specifically, $A$ is a $N_{R} \times 3$ vector. $V$ is a $M \times T_{0}$ vector, where $T_{0}$ is the total time in the dataset. $C$ is a $M \times 1$ vector. $R$ is a $N_{R} \times M$ vector. Our problem is defined as follows:}

\textbf{Problem (learning to route)} Given the road network $G$, real-time traffic $V$, and drivable conditions $C$, the goal is to learn a generative function $f$ that can produce the most possible routes $R$ for input $A$ (origin, departure time, and destination). Mathematically, the goal is to minimize the total loss over the data:

\begin{equation}
    \mathcal{L} = \sum_{i} l(R_{i}, f(A_{i},V[:,t=T_{i}], C, G))
    \label{loss}
\end{equation}
where $l$ is the loss function, and $f$ generates the routes (probabilities for choosing each road) for given $A$, $V$, and $C$. In our experiments, we use categorical cross-entropy as $l$.

Note that, our learning to route problem sets only one route for a given origin, destination, departure time, and real-time drivable condition. We admit that different people may take different routing decisions. \textbf{However, due to the lack of human-level feature data, we can not learn the routing decisions of every individual.} What we only can do is to learn one routing decision that most people choose, or to learn the proportion of different routing decisions.

%% file: method.tex
\section{Method}
\label{sec:method}







\subsection{Overview}

As we mentioned before, human routing decisions are determined by two major categories of factors: one is time or distance, and the other is the drivable conditions. Intuitively, we can directly learn a data-driven model that takes those factors as input to predict drivers' routes. However, in many cases, the route data is very limited due to issues such as device accuracy and privacy~\cite{lian2018one}. Thus, pure data-driven models can hardly get satisfactory results.

\nop{

As we mentioned before, human routing decisions are determined by many factors. Take commuting as an example. When people go to work in the morning, they will plan routes to arrive in their workplaces through the shortest or fastest way. However, for most of the cases, there are several candidate routes sharing similar total travel time. Under this circumstance, people will also judge a route by its drivable condition, such as speed limit, number of lanes, smoothness of road surface, and number of traffic lights. Actually, in the real data, only 50$\%$ of the total vehicle trajectories have followed the fastest route, which means the drivable condition plays a significant role when people choose their routes.

Hence, there are two major categories of factors affecting human routing decisions: one is time or distance, and the other is the drivable condition. Intuitively, we can directly learn a data-driven model that takes those factors as input to predict drivers' routes. However, in many cases, the route data is very limited due to issues such as equipment accuracy and privacy. Thus, this pure data-driven model can hardly get satisfactory results.}

To solve this problem, we propose to borrow knowledge from theoretical shortest path algorithms, such as Dijkstra and A* algorithms. We propose a residual framework to combine the theoretical part and the data-driven part, where the theoretical part is designed for finding the shortest path, and the data-driven part is designed for capturing the drivable condition preferences. We notice that the raw output of the neural network may not be a valid path. Therefore, we design a path validator to process the raw output and make it valid on the graph.
  
\begin{figure}[htbp]
\centering
\includegraphics[width=0.473\textwidth]{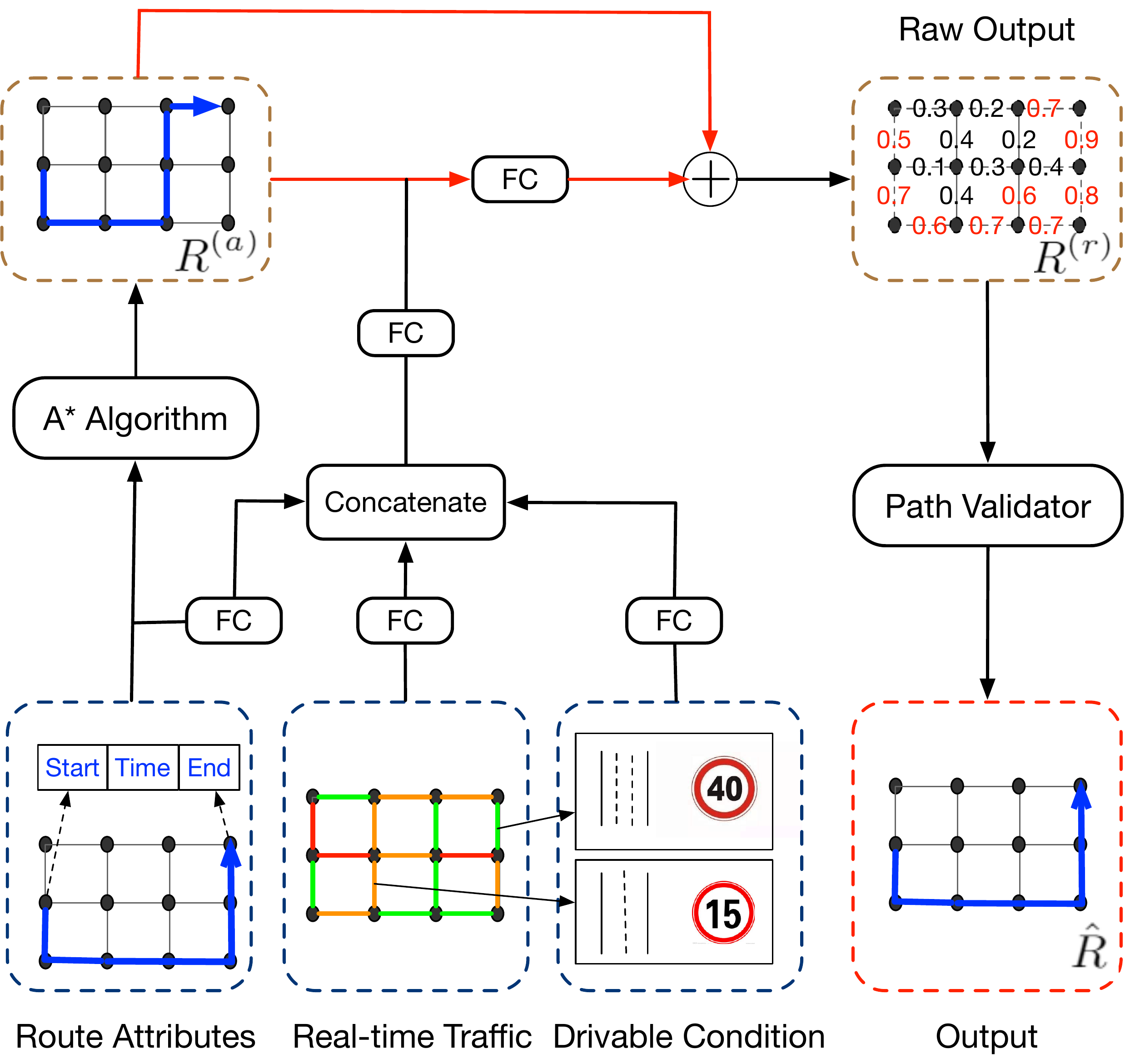}
\caption{Framework of \ours. The inputs of our model are route attributes $A$, real-time traffic $V$, and drivable condition $C$. The final output routes come in three steps: A* algorithm generates the fastest routes $R^{(a)}$; residual net generates the raw output routes $R^{(r)}$; path validator converts $R^{(r)}$ to the final output routes $\hat{R}$. Note the we use red lines to mark our residual learning part.}
\label{fig::model}
\end{figure}

In summary, our model contains three modules, theoretical algorithm, residual network, and path validator, called \ours (Learning to Route), as Figure~\ref{fig::model} shows. We will illustrate each of these modules in the following sections.

\subsection{Theoretical Algorithm}


To find the fastest path, we convert the focused region into a weighted graph, which has $N$ nodes (intersections) and $M$ edges (road segments). The weight on each edge is the travel time, calculated by real-time traffic $V$. 

We implement the A* algorithm as our theoretical part. A* algorithm uses a heuristic function $h$ to determine the direction in searching. This search strategy is very similar to the fact that humans can only observe the traffic of a few roads ahead when in estimating the travel time and choosing path correspondingly. We have also tried using the Dijkstra algorithm. It yields similar results but takes a much longer time to run. We believe it is because Dijkstra always tends to traverse all intersections, which is very time-consuming. In contrast, the searching strategy of A* algorithm can save much searching time in a grid-like city roadnet.

Equation~\eqref{A*} shows the fundamental idea of A*. Given current node $c$, and destination $e$, the score function value $F(x)$ for a candidate node $x$ can be calculated as Equation~\eqref{A*}, where $g(c,x)$ is the actual travel time from $c$ to $x$ and $h(x,e)$ is the estimated travel time from $x$ to $e$, calculated by the Euclidean distance~\cite{anton1994elementary} of $(x,e)$ and the highest limited speed. A* algorithm will choose the node with the highest score at each time step.
\begin{equation}
F(x) = g(c,x) + h(x,e)
\label{A*}    
\end{equation}
Finally, A* algorithm provides the fastest routes $R^{(a)}$.

\subsection{Theory-Guided Residual Structure}





To further incorporate the effect of drivable conditions, we design a theory-guided residual structure. We utilize the idea of residual learning in ResNet~\cite{he2016deep}. The inputs of our theory-guided residual network are: route attributes $A$, real-time traffic $V$, drivable condition $C$, and output from A* algorithm $R^{(a)}$. First, we use three embedding layers to process $A$, $V$, and $C$:
\begin{equation}
\begin{split}
 Z_{1} &= \sigma(W_{1} * A + b_{1}), \\
 Z_{2} &= \sigma(W_{2} * V[:,t = T] + b_{2}), \\
 Z_{3} &= \sigma(W_{3} * C + b_{3}), \\
\end{split}
\label{4.3.1}    
\end{equation}
where $W_{1}, W_{2}, W_{3}$ are weight matrices, $b_{1}, b_{2}, b_{3}$ are bias vectors, $\sigma$ is the activation function, $V[:,t = T]$ denotes the average speed on all roads at the same time interval as the departure time $T$ in $A$. 

Then, we concatenate the results after three embedding layers, and use one more embedding layer to process them:

\begin{equation}
\begin{split}
 Z_{4} &= \sigma(W_{4} * [Z_{1}, Z_{2}, Z_{3}] + b_{4}), \\
\end{split}
\label{4.3.2}    
\end{equation}
where $W_{4}$ is a weight matrix, $b_{4}$ is a bias vector.

Finally, we begin to utilize $R^{(a)}$, the output from A* algorithm. We concatenate $R^{(a)}$ and $Z_{4}$, and let them go through another fully connected layer. The fundamental idea of residual learning is to add the result with the origin $R^{(a)}$. By this way, we can get the raw output routes $R^{(r)}$:

\begin{equation}
\begin{split}
R^{(r)} &= \sigma(W_{5} * [R^{(a)}, Z_{4}] + b_{5}) + R^{(a)},
\end{split}
\label{4.3.3}    
\end{equation}
where $W_{5}$ is a weight metric, $b_{5}$ is a bias vector. The implementation details of our neural networks are described in Section Experiment. The raw output routes $R^{(r)}$ are used for loss calculation and optimize the whole network. 

It seems that the result of the theoretical part and the other part are simply combined by full connected networks. However, it contains the idea of residual learning in ResNet~\cite{he2016deep}, which is theoretically guaranteed.
 
\begin{figure}[ht!]
\centering
\begin{tabular}{cc}
\includegraphics[width=0.18\textwidth]{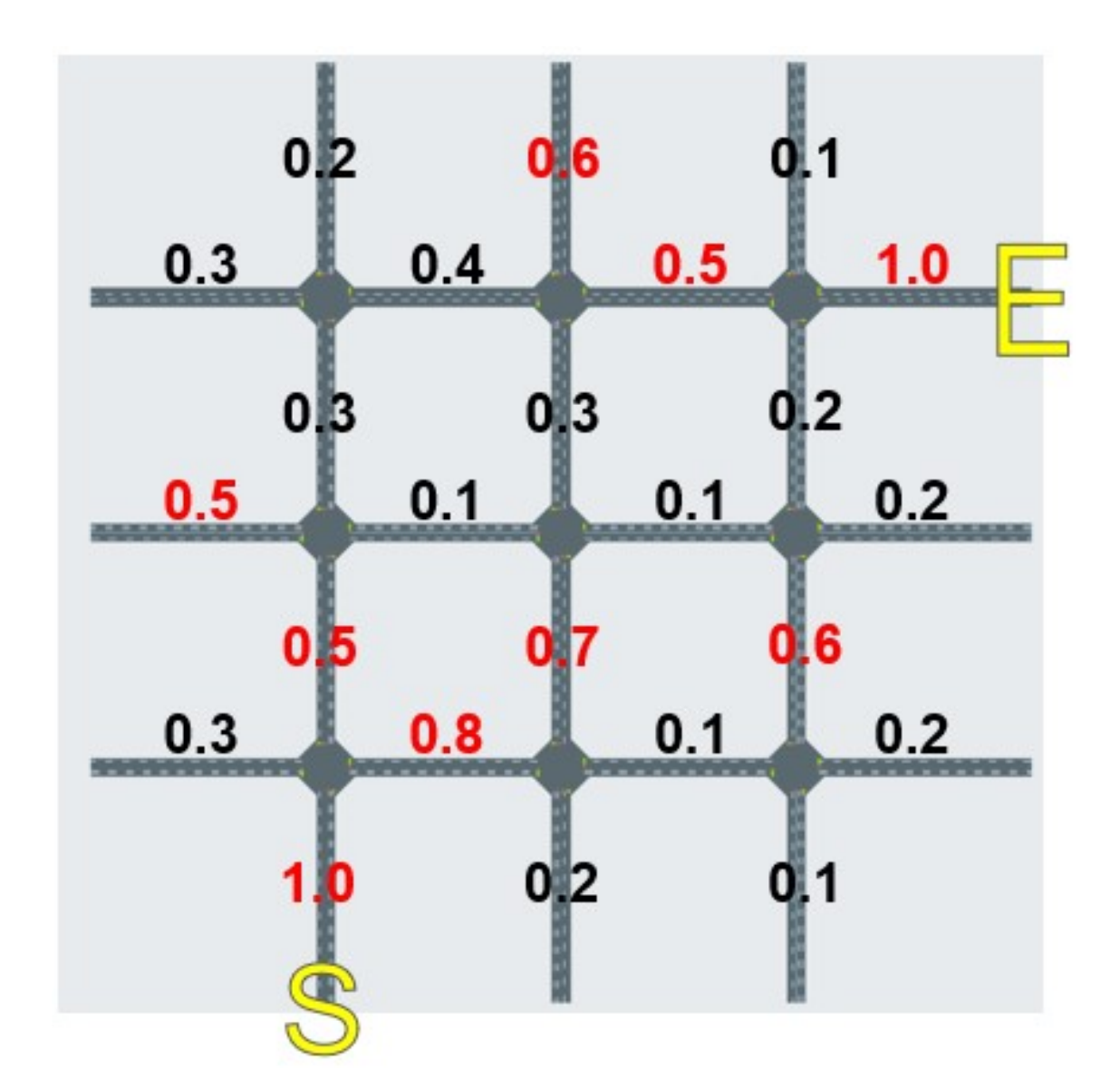} & 
\includegraphics[width=0.18\textwidth]{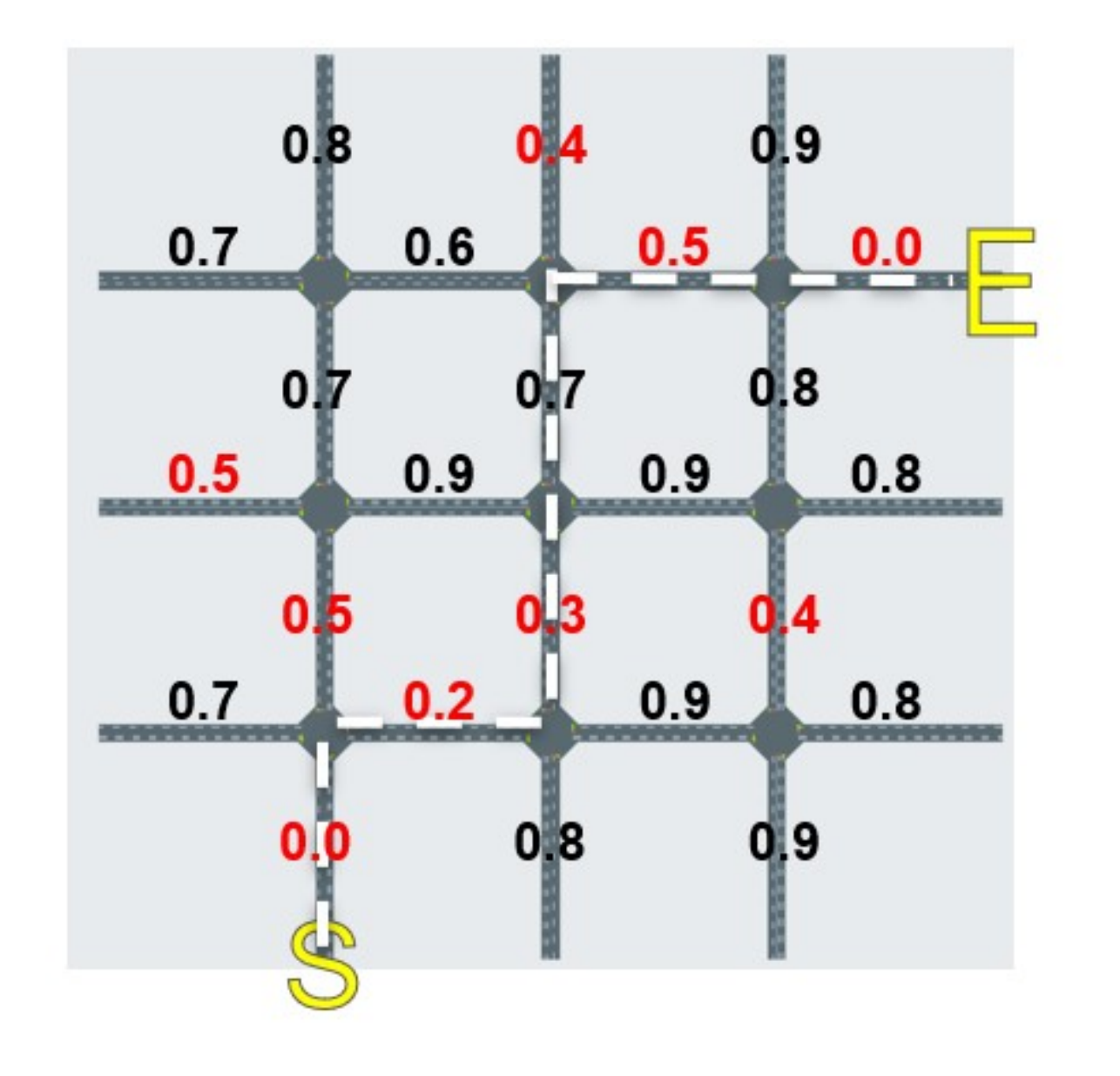} \\
(a) $R^{(r)}_{i}$ & (b) $1 - R^{(r)}_{i}$
\end{tabular}
\caption{The illustration of our path validator. Figure (a) denotes one of raw output routes $R^{(r)}_{i}$ from the residual network, and every $R^{(r)}_{i,j}$ is shown on $j$-th road. We mark the higher values as red, however, we can see the red numbers are not connected. In figure (b), we change the weight to $1 - R^{(r)}_{i,j}$. Then, we can find a shortest path in figure (b) as the white dotted line, which is a valid path.}
\label{fig::validator}
\end{figure}

\subsection{Path Validator}



After the residual net, we can get the raw output routes $R^{(r)}$, where $R^{(r)}_{i,j}$ denotes the predicted value of $j$-th road for the $i$-th route. The value of $R^{(r)}_{i,j}$ is number between 0 and 1, representing the probability that road segment $j$ is selected in route $i$. To convert $R^{(r)}_{i}$ to a binary vector, a straightforward way is to set a threshold. However, no matter how we set the threshold, $R^{(r)}_{i}$ may still be an invalid path on the graph. Taking Figure~\ref{fig::validator}(a) as an example, we use $0.5$ as threshold and use red to mark the higher value. As we can see, there is not a connected path from the start point to the end point. 

To solve this problem, we design a path validator to process the raw route $R^{(r)}_i$. We regard $R^{(r)}_{i,j}$ as the preference of $R^{(r)}_i$ over road segment $j$, where a larger probability value means a higher preference. Then, we only need to find a valid path from start point to end point with the highest total preference. As Figure~\ref{fig::validator}(b) shows, we first generate another graph $G'$, where:
\begin{equation}
    R'^{(r)}_{i,j} = 1 - R^{(r)}_{i,j}, \forall j
\end{equation}
\begin{table*}[htbp]
\centering
\caption{Overall performance comparison on real data w.r.t precision(P), recall(R), F1 score(F1), match distance(M), 90\% match(90\%). The proposed method \ours outperforms all the baselines. All measurements are the higher the better. We use F1 score to evaluate the best baseline.
} 
\label{tab:overall_performance_real} 
\begin{tabular}{llllll|lllll|lllll}
\toprule
       & \multicolumn{5}{c}{\hangzhou} & \multicolumn{5}{c}{\porto} &
     
       \multicolumn{5}{c}{\beijing}   \\
          \cline{2-16}
       & P        & R        & F1       & M & 90\%     & P             & R        & F1   & M & 90\%      & P        & R        & F1   & M & 90\%   \\ \midrule
\dijdistance   & 0.70 & 0.56 & 0.62  & 53\%  & 37\% & 0.67 & 0.51 & 0.58 & 52\% & 37\% & 0.48 & 0.43 & 0.45 & 44\% & 37\% \\
\dijtime 
& 0.71 & 0.57 & 0.63 & 55\% & 38\%
& 0.59 & 0.47 & 0.52 & 49\% & 34\% & 0.47 & 0.42 & 0.44 & 43\% & 37\% \\
\astar&  0.72 & 0.57 & 0.63 & 57\% & 39\% & 0.68 & 0.52 & 0.59 & 52\% & 38\% & 0.47 & 0.47 & 0.47 & 46\% & 36\%\\
\query & 0.59 & 0.55 & 0.57 & 53\% & 33\% & 0.67 & 0.77 & 0.67 & 69\% & 54\% & 0.52 & 0.57 & 0.54 & 60\% & 50\%\\
\nn & 0.74 & 0.59 & 0.66  & 58\% & 34\%  & 0.66 & 0.49 & 0.57 & 51\% & 35\% & 0.55 & 0.46 & 0.50 & 48\% & 41\% \\

\lfd & \textbf{0.75} & \textbf{0.62} & \textbf{0.68} & \textbf{57\%} & \textbf{37\%}  & 0.69 & 0.52 & 0.59 & 56\% & 38\% & 0.55 & 0.45 & 0.50 & 47\% & 40\% \\
\prr &0.72 & 0.55& 0.63& 60\% & 40\% &0.78 &0.57 & 0.66 & 60\% & 46\% & 0.54 & 0.50 & 0.52 & 51\% & 41\%\\
\mscr &0.75 &0.57 & 0.65& 56\% & 38\%  & 0.75 & 0.52 &0.61 &55\% &45\% &0.57& 0.47& 0.51& 49\% & 41\%\\
\rrr &0.62 & 0.57& 0.60 &56\% &35\% & \textbf{0.70} & \textbf{0.74} & \textbf{0.73} & \textbf{70\%} & \textbf{56\%} &0.53
 & 0.56& 0.54 & 60\% & 50\%\\
 \favour &0.73 &0.57 &0.64 & 60\% & 41\%  & 0.77 &0.59 &0.67 & 62\% &48\% & \textbf{0.66} & \textbf{0.54}    & \textbf{0.59} & \textbf{49\%} &\textbf{40\%}\\
  \midrule
\ours & \textbf{0.82} & \textbf{0.68} & \textbf{0.74} &\textbf{68\%}  & \textbf{45\%}  & \textbf{0.93} & \textbf{0.77} & \textbf{0.84} & \textbf{80\%} & \textbf{66\%} & \textbf{0.69} & \textbf{0.61} & \textbf{0.65} & \textbf{62\%} & \textbf{52\%}\\
\bottomrule  
\end{tabular}
\end{table*}
Hence, the shortest path in $G'$ will corresponds to the highest preference path in $G$. Therefore, we can use the shortest path algorithm to find the shortest path and output it as $\hat{R}$. Since $\hat{R}$ is the result of the shortest path algorithm, it must be valid. Note that, the path validator and the final output $\hat{R}$ are not for loss propagation. They are just to make the output more reasonable.

%% file: experiments.tex
\section{Experiment}
\label{sec:experiment}


\subsection{Research Questions}
\label{sec:experiment-rq}
In this section, we conduct experiments to answer the following questions.
\begin{itemize}
    \item RQ1: Compared with baseline methods, how is the proposed method \ours performing on learning human routing decisions?
    \item RQ2: How do these methods perform under different amount of data? 
    \item RQ3: What determines humans' real routing decisions? What are the reasons that make other baseline methods fail to predict the real routing decisions?
\end{itemize}


\subsection{Implementation}
For the input drivable condition for each road, we use the speed limit, the number of drivable lanes, road level, the width of each lane, and the number of passing bus lines as features. All these road features are collected from OpenStreetMap~\footnote{https://www.openstreetmap.org/}. We follow the implementation of \astar algorithm in~\cite{kim2005optimal} to build the theoretical part.
We have released our code and datasets in an anonymous repository~\footnote{https://bit.ly/3dK2MEa}. The detail information of hyperparameters can been found in this repository.


\subsection{Dataset}

\begin{figure}[htpb]
    \centering
    \begin{tabular}{ccc}
    \includegraphics[width=0.15\textwidth]{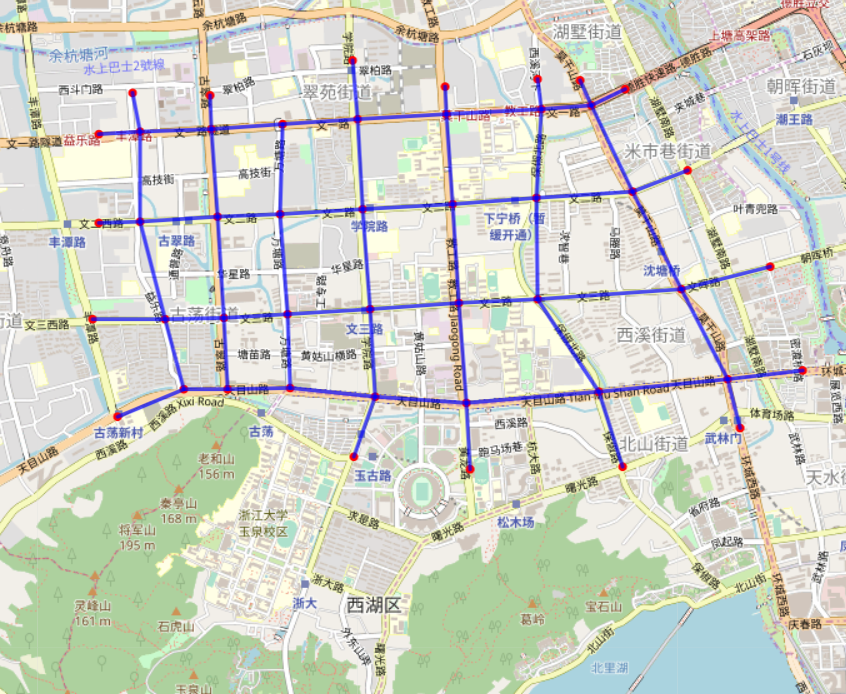} & 
    \includegraphics[width=0.134\textwidth]{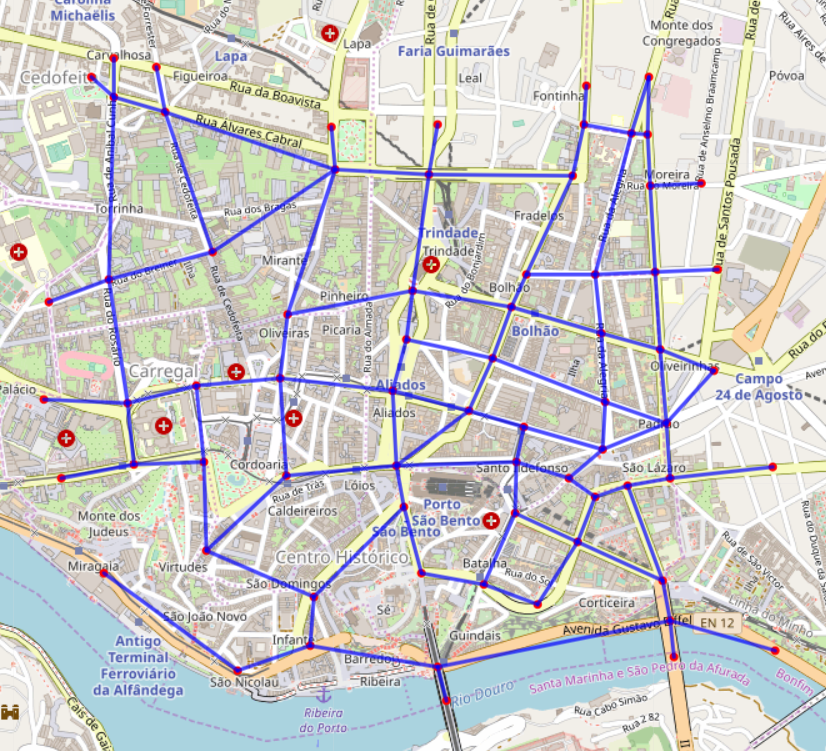} & 
     \includegraphics[width=0.134\textwidth]{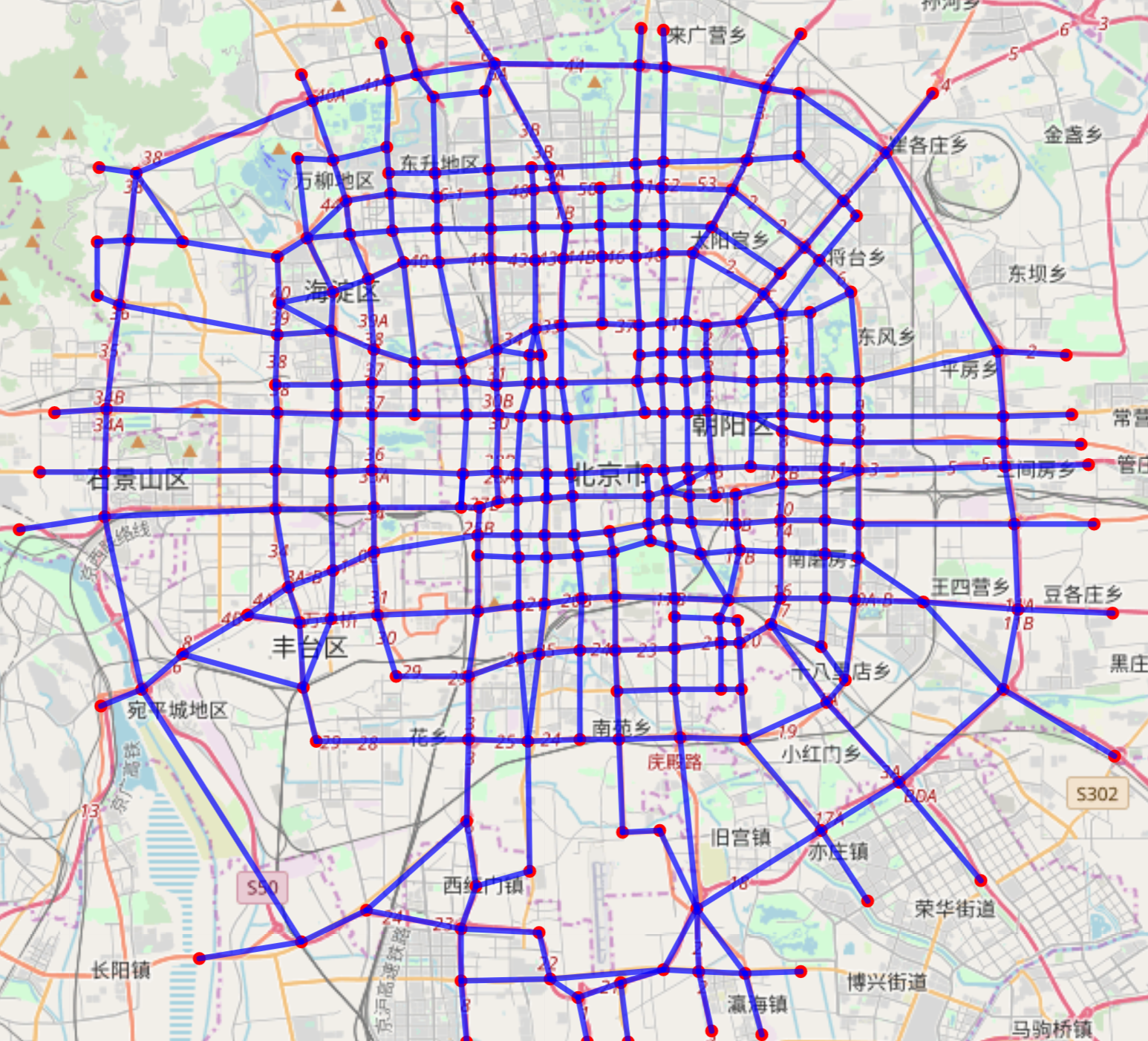} \\
     (a) Hangzhou & (b) Porto & (c) Beijing \\
    \end{tabular}
    \caption{The areas of the three real datasets.}
    \label{fig:bj_map}
    \end{figure}

We use one synthetic dataset and three real-world vehicle route datasets to conduct the experiments. The areas we select in real datasets are shown in Figure~\ref{fig:bj_map}. We obtain real-time traffic data from Google Maps and the road condition information from OpenStreetMap.
We randomly separate the data to train set (75\%), validation set (5\%), and test set (20\%). The input drivable condition data $C$ of all datasets are normalized 100-dimension vector. 

\begin{itemize}

    \item \textbf{Synthetic data}. This dataset is generated by an advanced traffic simulator Cityflow~\cite{zhang2019cityflow}. We use a 3x3 grid road network with more than 1000 different vehicles to run a four-hour simulation. 
    
    \item \textbf{Hangzhou data}. This dataset contains 9,656 one-day routes of taxis in Hangzhou, China on January 6, 2019. 
    We conduct experiments in the downtown area of Hangzhou, which contains 54 intersections and 63 roads.
    \item \textbf{Porto data}~\cite{moreira2013predicting}. This dataset provides 7,734 records of taxi orders from July 1, 2013, to June 30, 2014. 
    The experiment is conducted in the downtown area with 69 intersections and 100 roads.
    \nop{There are 69 intersections, 100 roads, and 2576 routes in this area.}
    \item \textbf{Beijing data}~\cite{lian2018one}. This dataset covers the taxi routes per day in May 2009, with about 160,000 records.
    All the main road segments inside Beijing's 5th ring road are taken into consideration, which yields 300 intersections and 522 road segments finally.

\end{itemize}



\begin{table*}[htpb]
\centering
\caption{Performance comparison under different amount of data w.r.t. F1 score, under synthetic dataset and real-world dataset (Beijing dataset). The last row ``Improvement'' shows the relative improvement over the best baseline. 
}\label{tab:overall_performance_sparsity_bj}

\begin{tabular}{llllll | lllll}
\toprule & \multicolumn{5}{c}{Real-World Dataset} & \multicolumn{5}{c}{Synthetic Dataset} \\
\cline{2-11}
      &   100\% & 50\% & 20\% & 5\% & 1\% &   100\% & 50\% & 20\% & 5\% & 1\%\\ \midrule
\dijdistance & 0.45 & 0.45 & 0.45 & 0.45 & 0.45& 0.41 & 0.41 & 0.41 & 0.41 & 0.41\\
\dijtime  & 0.44 & 0.44 & 0.44 & 0.44 & 0.44 & 0.42 & 0.42 & 0.42 & 0.42 & 0.42 \\
\astar  & 0.47 & 0.47 & 0.47 & 0.47 & 0.47  & 0.45 & 0.45 & 0.45 & 0.45 &  0.45\\
\nn &0.50 &0.49 &0.47 &0.45 &0.43 & 0.72 & 0.68 & 0.58 & 0.51 & 0.49\\

\lfd & 0.50 & 0.48 & 0.48 & 0.46 & 0.44 & \textbf{0.73} & 0.67 & 0.63 & 0.51 & 0.45\\
\prr &0.52 &0.51 &0.51 &0.50 & \textbf{0.49}  &0.71 &\textbf{0.68} &\textbf{0.64} &0.60 &0.55 \\
\mscr &0.51 &0.51 &0.50 &0.49 &0.48 &0.64 &0.62 &0.62 &\textbf{0.61} &\textbf{0.57} \\
\rrr &0.54 &0.54 &\textbf{0.53} &\textbf{0.51} & 0.48  &0.70 &0.67 &0.63 &0.60 &0.56 \\
\favour &\textbf{0.59} &\textbf{0.56} &0.51 &0.49 &0.47 &0.59 &0.59 &0.57 &0.56 &0.55\\
\midrule
\ours &\textbf{0.65} &\textbf{0.64} &\textbf{0.62} &\textbf{0.60} &\textbf{0.59} & \textbf{0.79} & \textbf{0.76} & \textbf{0.73} & \textbf{0.71} & \textbf{0.68} \\
\midrule
Improvement &10\% & 14\% & 17\% & 18\% & 20\% &8\% & 12\%& 14\%& 16\%&19\% \\
\bottomrule  
\end{tabular}
\end{table*}

\subsection{Compared Methods}

\begin{itemize}
    \item \textbf{\dijtime}~\cite{shimizu1995route}: This method converts the road network into a weighted graph, where the weights are the estimated travel time on roads, then runs the Dijkstra algorithm to find the shortest path. 
    \item \textbf{\dijdistance}~\cite{hamada2014route}: This method is similar to the previous one, except that this method uses distance directly as the edge weight. 
    \item \textbf{\astar}~\cite{kim2005optimal, bell2009hyperstar, nakajima2012route}: 
    This method adapts the original Dijkstra algorithm with a heuristic function (e.g., estimated distance to the destination) to determine the next node to search. 
    \item \textbf{\query}~\footnote{http://www.google.com/mobile/navigation/}:
    This method queries the most commonly used map apps, Gaode map (for \hangzhou and \beijing), and Google Maps (for \porto) for recommended routes, using origin, time and destination as input. 
    \item \textbf{\nn}~\cite{lecun1998gradient, jia2020context}:
    This method uses a neural network to predict routes given the current traffic situation. This network contains two fully connected layers, with ReLU as the activation function.
    \item \textbf{\lfd}~\cite{argall2009survey, chen2017pathrec} (Learning from Demonstration): This method first generates some data with the theoretical model and uses the combination of generated data and real data to train the neural networks. 
    \item \textbf{\prr}~\cite{dai2015personalized, Ghagage2018ASO}: Based on people's historical choice of routes, this method first predicts users’ preferences in the distance, time, and fuel consumption. Then, it uses the learned preferences to recommend routes by linear regression.
    \item \textbf{\mscr}~\cite{dong2014recommend, Qu2020ProfitableTT}: This method assigns scores to each road segment based on historical data and then runs the Dijkstra algorithm to find the best path for the given origin and destination. 
    \item \textbf{\rrr}~\cite{wang2014r3, santos2018context}: This method integrates historical taxi driving data and real-time data to provide users with route recommendations, which will make a balance on roads to alleviate congestion. 
    \item \textbf{\favour}~\cite{campigotto2016personalized}: This method proposes to learn users' profiles from the routes chosen by them in history.
    The profiles of users take users' routing preference and traveling frequency into consideration.
    
    \end{itemize}

\subsection{Metrics} 

    We use five frequently used metrics to evaluate the results, which are widely accepted in the previous work~\cite{ziebart2008maximum, Ghagage2018ASO}: \textbf{Precision} is the ratio of correctly predicted positive roads to the total predicted positive roads. \textbf{Recall} is the ratio of correctly predicted positive roads to all positive roads in the real routes. \textbf{F1 score} is the weighted average of Precision and Recall. \textbf{Match distance ratio} is the ratio of the total distance of correctly predicted selected roads to the total distance of real selected roads. \textbf{90\% match}~\cite{ziebart2008maximum} measures the portion of the predicted routes that match the real routes with at least 90\%.

\begin{figure*}[htpb]
    \centering
    \begin{tabular}{ccc}
   \includegraphics[width=0.3\textwidth]{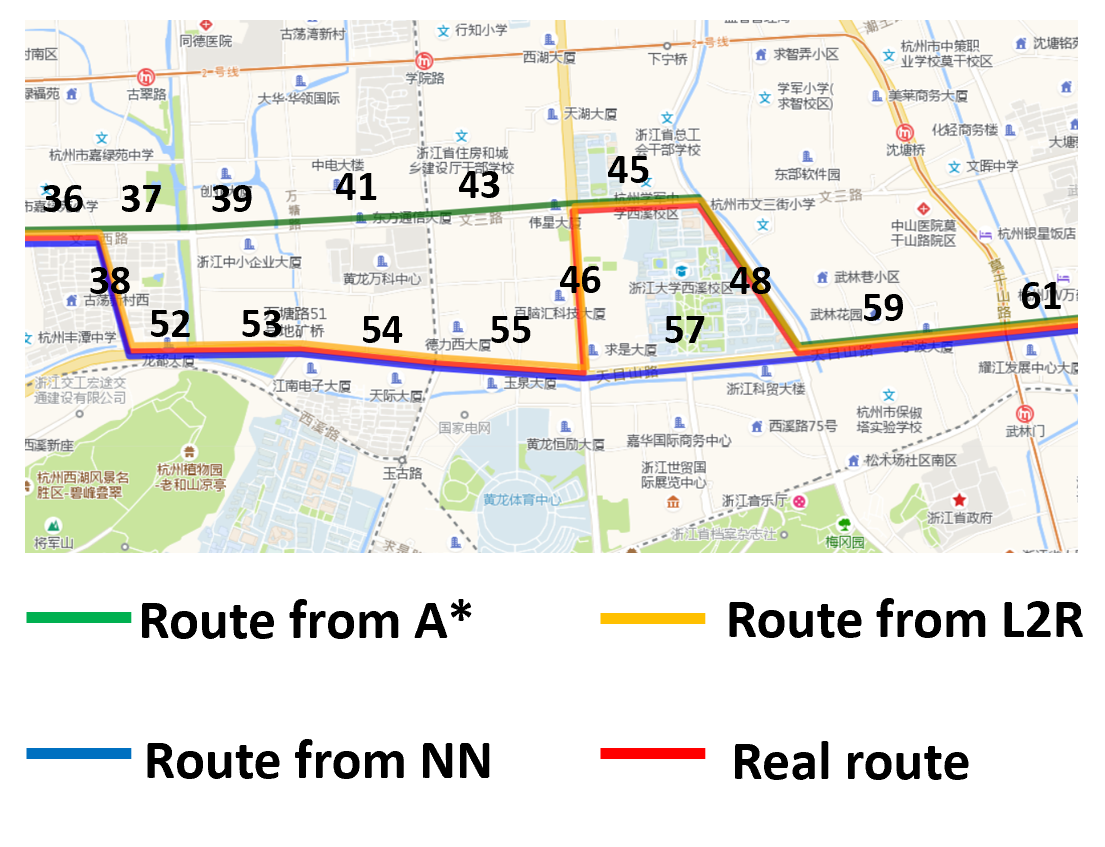} &
    \includegraphics[width=0.3\textwidth]{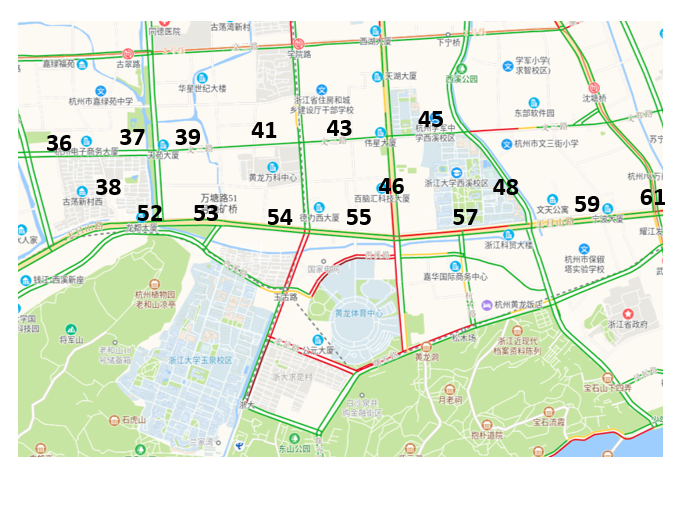} &
    \includegraphics[width=0.3\textwidth]{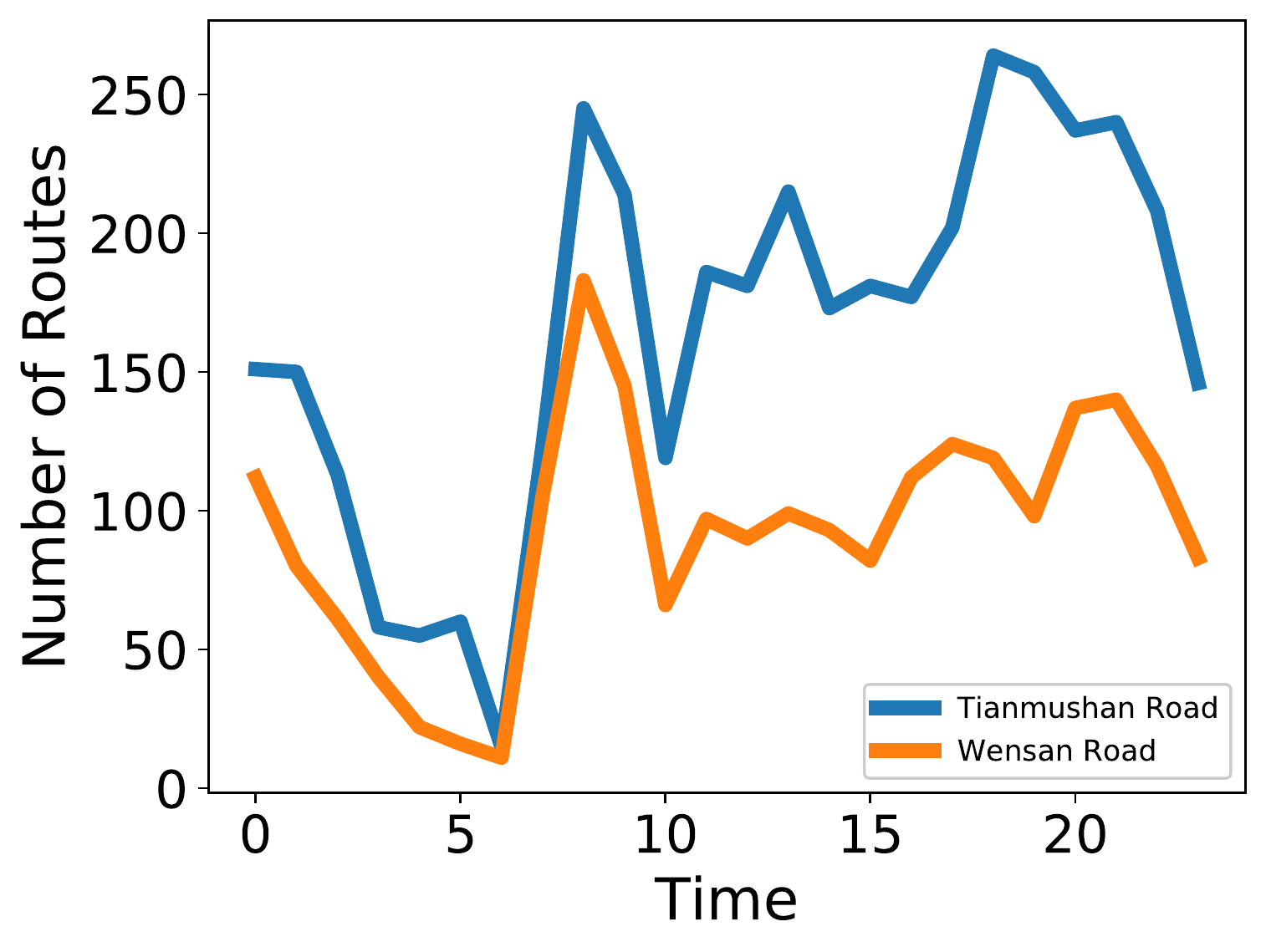}\\ 
    (a) Four routes by different model & 
    (b) Real-time traffic & (c) Traffic volume on two roads\\
    \end{tabular}
    \caption{Case Study: Recovered routes from a shopping center (road No. 61) to a residential area (road No. 36) in Hangzhou. (a) shows routes from different models and the real route. (b) shows the real-time traffic conditions in this area, where ``green'' represents light traffic while ``red'' stands for heavy congestion. Road No.57 is heavily congested. (c) shows traffic volume on Tianmushan Road (61-59-57-55-54-53-52) and Wensan Road (45-43-41-39-37-36) in a day, where Tianmushan Road always has a larger traffic volume than Wensan Road.}
    \label{fig:hangzhou_case}
\end{figure*}

\subsection{RQ1: Comparison with Baseline Methods}
We compare our model and other baselines in both the synthetic dataset and real datasets. The results are shown in Table~\ref{tab:overall_performance_real}. (Due to the page limit, we do not show the overall performance on the synthetic dataset.) It turns out that \ours performs the best in all datasets and metrics. \rrr and \favour also achieve good performance under some metrics, since they both utilize and analyze historical data to extract portraits of users or roads. The performances of \prr and \mscr do not consistently perform well on all the metrics. We believe the reason is that the preferences of users or scores on roads assigned by their model can be biased since they highly rely on the historical data. The performance of \lfd surpasses \nn a little bit, for that \lfd has further utilized the power of theoretical models compared to \nn. The performance of \query is unstable because it depends on the accuracy of the map application in different cities. \dijdistance, \dijtime, and \astar perform poorly on all three datasets, which is reasonable since they rely purely on the distance and travel time estimation and ignore user preference. 

\subsection{RQ2: Performance comparison under different amount of training data}
In order to investigate how the performance will vary with different amounts of data, we conduct experiments with varying percentages of training data (100\%, 50\%, 20\%, and 1\%) on both the synthetic dataset and the real-world dataset. 

The results are shown in Table~\ref{tab:overall_performance_sparsity_bj}. With the decreasing amount of training data, the performances of all the methods tend to downgrade, while \ours maintains decent performance even with 1\% data. Though the results of \favour and \lfd are close to the result of our method when the data is sufficient, their performance will drop significantly when we reduce the training data size. Hence, the relative improvement of \ours over other methods are more significant when less data is available. 
Therefore, our method \ours can significantly alleviate the insufficient data problem.

\nop{Our model performs the best in all data scales. We can see that though \nn, \astarnn, and \lfd are close to our method when the data is sufficient, their performance will drop significantly when we reduce the data. The improvement between our model and the best baseline is increasing as the data become sparse, which illustrates that our model is less sensitive to data sparsity. The only exception is our performance on the Beijing case with 1\% data is worse than the one with 5\% data. This is because, in those two cases, the best baseline is \query, which is not influenced by the data sparsity.
}

\subsection{RQ3: Case Study on Recovered Routes}

We take a closer look at a frequent trip from the Wulin shopping center to a residential area in Hangzhou. We select routes of the drivers with the same departure time, origin, and destination. The routes predicted by different methods and the real route are shown in Figure~\ref{fig:hangzhou_case}(a). We can see that neither \nn nor \astar can fit the real route, while \ours predicts a route the same as the real route. 
\nop{ Next, We will analyze each route:
Actually, the real route is like a combination of \nn route and \astar route.}

For \astar (green route), this method will predict the fastest route according to the real-time traffic situation. As we can see in Figure~\ref{fig:hangzhou_case}(b), there is heavy congestion in road No.57. Therefore, \astar decides to avoid road No.57 and detour by road No.48 to Wensan Road (45-43-41-39-37-36). 

For \nn (blue route), this method learns from historical data. It has a high probability to choose to drive on Tianmushan Road (61-59-57-55-54-53-52), since more drivers in historical data choose Tianmushan Road than Wensan Road, as Figure~\ref{fig:hangzhou_case} (c) shows. It can be explained by their descriptions in OpenStreetMap: Tianmushan Road is a primary highway, which has 8 to 10 lanes with a 70km/h speed limit; Wensan Road is a secondary highway, which has 4 lanes with a 40km/h speed limit. Therefore, people prefer to drive on Tianmushan Road for a better drivable condition.

For the real route (red route), drivers realize that road No.57 is heavily congested, so they decide to detour as \astar predicted. However, due to the lower speed limit and few drivable lanes on Wensan Road, they choose to detour again back to Tianmushan Road by road No.46. Finally, they will drive on Tianmushan Road to the destination, as \nn predicts. Since \ours is a combination of the theoretical algorithm and neural network, \ours successfully recovers the same route (orange route) as the real route. 

Indeed, there are still some drivers that do not follow the detour route as \ours predict. According to our statistic on the route data in the dataset, for the drivers that have the same origin, destination, and departure time, about 80\% drivers follow the orange route, 10\% drivers follow the blue route, 5\% drivers follow the green route, and last 5\% drivers follow other routes. Based on the statistical results, we can see that most of the drivers follow the route predicted by \ours. 

\subsection{Time Efficiency Analysis}
In the real-time route recommendation problem, the time complexity is an important metric.
However, in learning to route problem, time complexity is not important, since our training can be done offline, and the testing takes little time. 

During the experiments, we find that the most time-consuming part is the theoretical part, the \astar algorithm. Since we have enough time for offline training, the online testing time used for the neural network is not significant compared to \astar algorithm.
If we consider the worst case, the time complexity of \ours is $O(N \times log(N) + M)$.






%% file: conclusion.tex
\section{Conclusion}
\label{sec:conclusion}


In this paper, we propose to learn human routing decisions from human mobility data.
We design a residual learning framework \ours to combine a data-driven model and classic routing algorithms, where we can take advantage of the neural nets to learn the multiple impacting factors while relying on \astar algorithm to alleviate the insufficient data problem. In the extensive experiments, our proposed method has shown its ability in delivering strong performance, especially under limited data. For future work, we will focus on how to transfer the learned model from one city to a new city with less or no data, which will make our model more applicable.

\section*{Acknowledgment}
This work was supported in part by NSF awards \#1652525, \#1618448 and \#1639150. The views and conclusions contained in this paper are
those of the authors and should not be interpreted as representing
any funding agencies.

%% file: reference.tex
{\small
\bibliographystyle{ieee_fullname}
\bibliography{egbib}
}